\documentclass[pdflatex,sn-vancouver-ay]{sn-jnl}
\usepackage{graphicx}%
\usepackage{multirow}%
\usepackage{amsmath,amssymb,amsfonts}%
\usepackage{amsthm}%
\usepackage{mathrsfs}%
\usepackage[title]{appendix}%
\usepackage{xcolor}%
\usepackage{textcomp}%
\usepackage{manyfoot}%
\usepackage{booktabs}%

\usepackage[linesnumbered,ruled,vlined]{algorithm2e}
\usepackage{listings}
\usepackage{array}
\usepackage{hyperref}

\usepackage{bigstrut} 
\usepackage{float}
\usepackage{tabularx}
\newcommand{\virgolette}[1]{``#1''}
\usepackage{bm}
\usepackage{caption}
\usepackage{subcaption}

\theoremstyle{thmstyleone}

\theoremstyle{thmstyletwo}

\theoremstyle{thmstylethree}

\usepackage[ruled,vlined]{algorithm2e}
\usepackage{amsmath,amssymb}
\usepackage{bm}
\usepackage{needspace}

\begin{document}

\title[Simultaneous Latent Budget Trees for Stratified Classification]{Simultaneous Latent Budget Trees for Stratified Classification}

\author[1]{\fnm{Cristian} \sur{Buoncompagni}}\email{cristian.buoncompagni@unina.it}
\author[1]{\fnm{Stefano} \sur{Pellegrino}}\email{stefano.pellegrino@unina.it}
\author*[2]{\fnm{Giulia} \sur{Vannucci}}\email{giulia.vannucci@unina.it}
\author[3]{\fnm{Raffaele} \sur{Dubbioso}}\email{raffaele.dubbioso@unina.it}
\author[2]{\fnm{Roberta} \sur{Siciliano}}\email{roberta.siciliano@unina.it}

\affil*[1]{\orgdiv{Department of Physics}, \orgname{University of Naples Federico II}, \orgaddress{\street{via Cintia 21}, \city{Napoli}, \postcode{80125}, \country{Italy}}}
\affil*[2]{\orgdiv{Department of Electrical Engineering and Information Technologies}, \orgname{University of Naples Federico II}, \orgaddress{\street{via Claudio 21}, \city{Napoli}, \postcode{80125}, \country{Italy}}}
\affil*[3]{\orgdiv{Department of Neuroscience, Reproductive Sciences and Odontostomatology}, \orgname{University of Naples Federico II}, \orgaddress{\street{via Sergio Pansini 5}, \city{Napoli}, \postcode{80131}, \country{Italy}}}

\abstract{In the era of Explainable Artificial Intelligence, there is a renewed focus on single trees for their ease of interpretation. This paper introduces Simultaneous Latent Budget Trees, a probabilistic machine learning framework for classification trees in the presence of a stratification factor such as a temporal, spatial, or demographic variable, acting as a control variable or potential confounder. Standard tree growth procedures are not designed to optimize a conditional split rule. A model-based split rule is proposed in which child nodes are interpreted as latent components of a simultaneous mixture model, such as the Simultaneous Latent Budget Model and its constrained versions, fitted to the parent node. Mixing parameters drive the observations, differently for each group, to the child nodes whereas latent budgets parameters update the response classes profile of each level of the control variable. Parameters are estimated by least squares considering a neural network perspective of the model. An informative tree structure can be interactively visualized with interpretation aids on the node and the paths, including visual pruning and decision tree selection procedure. Suitable measures are proposed to handle an unbalanced response class distribution. The proposed methodology is applied to investigate gender-related differences in disease progression of Amyotrophic Lateral Sclerosis. The SLBT library with the various tree-based algorithms is available in the linked GitHub repository. 
}

\keywords{Classification trees, Multiclass and Unbalanced Response, Simultaneous Latent Budget Model, Predictability Improvement, Visual Pruning}

\maketitle

\section{Introduction}

Classification trees represent one of the most robust frameworks for supervised machine learning, offering a unique balance between predictive power and structural interpretability.

The CART benchmarking methodology \citep{breiman1984} laid the foundation for numerous subsequent approaches \citep{mola1992, siciliano2000, hothorn2006, dusseldorp2004, dusseldorp2010, conversano2011, conversano2017, grubinger2014, dambrosio2016, kindo2016, fokkema2021, kim2022, vannucci2023} and remains a cornerstone to this day.

Let $(Y,{\bf X})$ be a multivariate random variable, where the response variable $Y$ takes values in the set of classes $j \in {\cal J}$, and ${\bf X}$ is the $M$-dimensional vector of explanatory variables or predictors with a numerical, ordinal, or nominal scale of measurement. A classification tree is built using a recursive binary segmentation of a learning sample ${\cal L} = \{(y_n, {\bf x}_n), n \in {\cal N}\}$ taken from $(Y, {\bf X})$ to reduce the impurity or heterogeneity of $Y$ due to predictors. Any node $\{t\}$ is a segment of the data matrix with $p(t)$ proportion of cases. In CART, the best split of any predictor is selected such to maximize the decrease of heterogeneity (or impurity) of $Y$ when passing from the {\it parent node} $\{t\}$ to the {\it left child node} $\{2t\}$ and the {\it right child node} $\{2t+1\}$\footnote{The child nodes of node $\{t\}$ are numbered $\{2t\}$ and $\{2t+1\}$ so that it is possible to uniquely associate to the tree graph a summary table describing the information of each node (i.e., the node $\{14\}$ is the left child node of node $\{7\}$ which is the right child node of node $\{3\}$ which is the right child node of node $\{1\}$).}. A stopping rule declares a node terminal, assigning a response label class. The set ${\tilde T}$ of leaves of the tree ${\cal T}$ forms the final partition with $\sum_{h \in {\tilde T}} p(h) = 1$. Cost-complexity pruning and cross-validation can be considered to identify the honest tree size of the decision tree for generalization to fresh data \citep{vapnik1995, vapnik1998}.

To improve predictive precision, researchers moved from single trees to ensemble methods such as bagging, boosting, and random forests \citep{breiman1996, breiman1998, friedman2000, breiman2001, hastie2009, zhou2012}. Although these methods offer superior accuracy, they often sacrifice interpretability, operating as black boxes where understanding the decision-making process is challenging. In contrast, algorithms based on single trees are valued precisely for their ease of interpretation, an increasingly crucial aspect in the era of Explainable AI (XAI). 

This renewed focus on single trees for XAI is the light motive of the methodological proposal of \emph{Simultaneous Latent Budget Trees} (SLBT) for classification in the presence of a stratification factor $Z$ such as a temporal, spatial, or demographic variable, acting as a control variable or potential confounder (Section \ref{sec:SLBT}). For this aim, performing separate trees using CART-like methods, each level of the stratification factor is an independent universe. Adding the stratification factor to the set of predictors does not provide information on how the explainable ability of a predictor $X$ acts on the response $Y$, conditional on $Z$, at each node $\{t\}$ of the tree. A third option might be to consider all predictors as compound variables of type $X \cdot Z$ resulting in a multiple split variable with the stratification variable at each node. This option is not only computationally intensive (as at each node a very large number of split variables are generated by all compound predictors) but the factor $Z$ would not act as a control variable (as the split rule does not optimize the explainable ability of the predictor $X$ {\it conditional to} $Z$ to predict the classification of the response profile for each level of $Z$).

SLBT can be understood as an extension of the Latent Budget Tree (LBT) \citep{siciliano1999}, recently applied in the field of medical robotics \citep{cotugno2025}, to the structure of the data cube with a set of learning samples, one for each level of $Z$. A model-based split rule is considered with a fast algorithm to identify the best split in terms of its ability to predict $Y$ conditional on $Z$. The name of SLBT derives from the model that governs the split, the Simultaneous Latent Budget Model (SLBM)\footnote{SLBM is an extension of the Latent Budget Model, originally introduced for the analysis of time budgets and then for any compositional data tables \citep{deleeuw1988, vanderHeijden1989, mooijaart1999, Jelihovschi2018}.}\citep{siciliano1994, Tambrea1999}. The key idea is that child nodes are understood as two latent components of a simultaneous mixture model fitted to the parent node. In the mixture, there are two sets of parameters; one drives the observations, differently for each group, to the child nodes, and the other updates the response profile of each level of $Z$ within the child nodes. Across-group homogeneity constraints on mixing parameters and/or latent budget parameters provide various options for the split rule. Parameters are estimated by least squares considering a neural network perspective of the model (Section \ref{sec:LS}). An informative tree-structure can be interactively visualized with interpretation aids on the nodes and the paths, including visual pruning and decision tree selection. SLBT algorithms have been applied in the study of Amyotrophic Lateral Sclerosis to investigate gender-related differences in disease progression (Section \ref{sec:ALS}). In the literature, several tree-based methodologies have been applied to the same disease, such as the GUIDE approach \citep{loh2020} and Model-Based Recursive Partitioning \citep{seibold2016}. Specifically, these identify subgroups of patients with heterogeneous therapeutic effects on functional decline, measured using the ALS Functional Rating Scale, and on survival time. In contrast to traditional recursive partitioning, SLBT embeds the stratification factor directly into the split logic via a latent budget framework, allowing simultaneous modeling of clinical predictors and subgroup-specific response profiles. The proposed methodology has been implemented in the SLBT library in Python and  C++ programming available in the linked GitHub repository (Appendix \ref{app:lib})\footnote{The details of the functions are reported in the package documentation available on the GitHub repository at \url{https://github.com/Giugurtah/SLBT}}. The pseudocodes of the algorithms for tree growth and visual tree selection are provided in the Appendix \ref{app:alg}.
As a result, the proposed SLBT methodology provides alternative algorithms with constrained and unconstrained model parameters in the split rule and an informative and interactively visualized tree structure for a complete simultaneous tree-based analysis of stratified classification (Section \ref{sec:con}).

\section{Simultaneous Latent Budget Trees}
\label{sec:SLBT}

\emph{Simultaneous Latent Budget Trees} (SLBT) is an innovative probabilistic machine learning framework for tree growth with a stratified classification in the presence of a control variable or potential confounder. SLBT predicts the response class of $Y$ based on the explainable ability of $M$ predictors under the control factor $Z$, taking values from the set $g \in {\cal G}$. The tree growth considers a data cube structure with a learning sample ${\cal L}_g = \{(y_{n_g}, {\bf x}_{n_g}), n_g \in {\cal N}_g\}$ taken from $(Y, {\bf X})$ for each level of $Z$. Predictors are categorical or categorized\footnote{In the SLBT library, a categorization algorithm has been implemented using $k$-means clustering with the Elbow method to determine the number of categories.}. The key issue is an explainable simultaneous mixture model-based split criterion with a fast algorithm to select the best split in terms of its ability to predict $Y$ conditional on $Z$\footnote{To reinforce the explanatory ability of the tree-growing method, it is worth categorizing numerical predictors to allow all predictors to play the same role in the analysis. In fact, continuous predictors may generate a larger number of split variables compared to categorical ones. Furthermore, in open source libraries nominal predictors are often treated as ordinal ones so that not all split variables are considered.}.

In the following, the methodological steps of SLBT are described in detail. The pseudo-codes of the algorithms for the tree growth and visual tree selection are provided in the Appendix (\ref{app:alg}).

\subsection{Ranking of predictors}
\label{subsec:varsel}

Predictors available at node $\{t\}$ are ranked in descending order by the partial predictability index $\tau$ of Gray and Williams for a set of two-way contingency tables\footnote{Consider the Gini index of heterogeneity of $Y$ and its ANOVA decomposition due a predictor $X$ into explained and residual components. The ratio between explained heterogeneity and total heterogeneity provides the Goodman and Kruskal predictability index $\tau$ \citep{goodman1954}, which is also associated with the CATANOVA testing procedure for two-way contingency tables \citep{light1971}. The extension to the set of two-way tables yields the multiple and partial predictability indices $\tau$ of Gray and Williams \citep{gray1975} related to the simple index by the following relation $\tau_{Y|X,Z} = \tau_{Y|X \cdot Z} - \tau_{Y|Z}$, where $\tau_{Y|X,Z}$ is the partial $\tau$ for the set of $G$ two-way tables $I \times J$, $\tau_{Y|X \cdot Z}$ is the multiple $\tau$ for the two-way table $I \cdot G \times J$ using the compound predictor $X \cdot Z$ and $\tau_{Y|Z}$ the simple $\tau$ for table $G \times J$. All indices $\tau$ range in $[0,1]$. It holds that $\tau_{Y|X,Z} = 0$ under conditional independence and $\tau_{Y|X \cdot Z} = 0$ under independence. Maximizing the partial index $\tau$ or the multiple index $\tau$ yields the same ranking of the predictors.}. For any predictor $X$ with categories in the set $i \in {\cal I}$, the partial index $\tau$ is defined as

\begin{equation}
\label{eq:taup}
        \tau_{Y|X,Z}(t) = \frac{\sum_i \sum_j \sum_g {(p_{j|i(g)}(t) - p_{j|+(g)}(t))}^2 p_{i+(g)}} {1 - \sum_j \sum_g p^2_{j|+(g)}(t)p_{++(g)}(t)}
    \end{equation}

\noindent
ranging in $[0,1]$, with conditional proportions $p_{j|i(g)}(t) = \frac{p_{ij(g)}}{p_{i+(g)}}$ and $p_{j|+(g)} = \frac{p_{+j(g)}}{p_{++(g)}}$, and margins $p_{++(g)} = \sum_i\sum_j p_{ij(g)}$. The (\ref{eq:taup}) evaluates the percentage improvement in predictability of $Y$ due to the predictor $X$ within each level of $Z$. 

The (\ref{eq:taup}) is equivalent to evaluating at node $\{t\}$ the proportional reduction in impurity of $Y$ due to the predictor $X$ conditional on $Z$ as: 

\begin{equation}
\label{eq:gamma}
    \gamma_{Y|X,Z}(t) = \frac{\sum_g i_{Y|(g)}(t)p_{++(g)(t)} - \sum_g \sum_i i_{Y|i(g)}(t)p_{i+(g)}(t)}{\sum_g i_{Y|(g)}(t)p_{++(g)}(t)}
\end{equation}

\noindent
where $i_{Y|(g)}(t) = 1 - \sum_j p^2_{j|+(g)}(t)$ and $i_{Y|i(g)}(t) = 1 - \sum_j p^2_{j|i(g)}(t)$ are the impurity measures of $Y$ within the $g$-th group, and within the $i$-th predictor category and the $g$-th group, respectively. Using the Gini index of heterogeneity as an impurity measure, (\ref{eq:gamma}) yields (\ref{eq:taup}). Alternative impurity measures such as the error rate or the entropy index can also be considered.

\subsection{Simultaneous mixture model-based split rule}
\label{subsec:SLBM}

The Simultaneous Latent Budget Model (SLBM) governs the model-based split rule. 
At node $\{t\}$, consider the current predictor $X$ with categories in the set $i \in {\cal I}$, the response variable $Y$ with classes in the set $j \in {\cal J}$ and the factor $Z$ with levels $g \in {\cal G}$. Let ${\bf p}_{Y|i(g)}'(t)$ be the $i$-th observed budget ($J$-dimensional row vector) of conditional proportions $p_{j|i(g)}(t)$ in the $g$th group of $Z$ summing up to one. SLBM approximates the observed budget by the theoretical budget, the $J$-dimensional (row) vector $\pi_{Y|i(g)}'(t)$, defined as a mixture of $K$ latent budgets ${\bf \beta}_{Y|k(g)}'(t)$, which are $J$-dimensional row vectors of elements $\beta_{j|k(g)}(t)$, with weights the mixing parameters $\alpha_{k|i(g)}(t)$:

\begin{equation}
\label{eq:SLBM}
    {\bf \pi}_{Y|i(g)}'(t) = \sum_{k=1}^K \alpha_{k|i(g)}(t) {\bf \beta}_{Y|k(g)}'(t) 
\end{equation}

\noindent
where $\alpha_{k|i(g)}(t)$ and $\beta_{j|k(g)}(t)$ are conditional probabilities, thus non-negative, and with $\sum_k \alpha_{k|i(g)}(t) = 1$ and $\sum_j \beta_{j|k(g)}(t) = 1$. (\ref{eq:SLBM}) is the unconstrained formulation in which the sets of model parameters are allowed to vary from table to table. Homogeneity constraints across-groups upon mixing and/or latent budget parameters can be considered\footnote{For $K = 1$, (\ref{eq:SLBM}) results in the conditional independence model $\pi_{j|+(g)}(t)$ and, in case of homogeneous latent budgets, it reduces to the independence model $\pi_{j|++}(t)$.}.

\begin{figure}[h]
    \centering
    \includegraphics[width=1\linewidth]{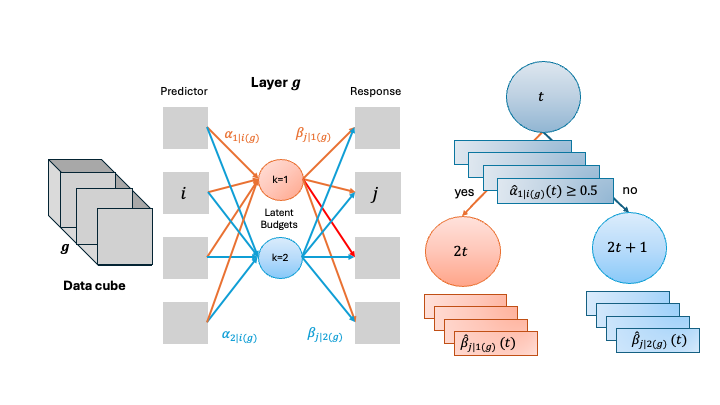}
    \caption{Simultaneous Latent Budget Tree: the neural network perspective of the model associated to the split rule}
    \label{fig: SLBT Split}
\end{figure}

In the binary tree, the SLBM, in the unconstrained or constrained version, with $K = 2$ fits to the data using the current best predictor. Two latent budgets are associated with the left and right child nodes. The model parameters are estimated using a suitable least-squares algorithm within the neural network formulation described in Section \ref{sec:LS}. Specifically, the mixing parameters connect the input layer (predictor) to the hidden layer, while the latent budget parameters link the hidden layer to the output layer (response). Figure \ref{fig: SLBT Split} describes the data cube, the neural network perspective of the model associated with the split rule.

The conditional probability estimate to belong to each latent class drives any observation to the left child or the right child node. This can be different for observations belonging to different groups of $Z$. Specifically, the split rule drives all cases in the $g$-th group with mixing parameter estimate ${\hat \alpha}_{1|i(g)}(t) \geq 0.5$ to the left child node $\{2t\}$, the others to the right node $\{2t+1\}$. Latent budget parameter estimates provide the prediction of the {\it response classification profile} for each level of $Z$ in the child nodes. 

\subsection{Split evaluation}
\label{subsec:splitev}

There is a fundamental step of split evaluation. A property of the $\tau$ index (\ref{eq:taup}) (and of \ref{eq:gamma})) is that $\tau_{Y|s,Z} \leq \tau_{Y|X,Z}$ for any split $s$ of the predictor $X$ that has generated it\footnote{This extends the result for the simple index $\tau$ \citep{mola1997, siciliano2000}.}, the former expresses the {\it Predictability Improvement} (PI) due to the split and the latter is its upper bound, the {\it Global Predictability Improvement} (GPI) due to its predictor. Both express an {\it information gain}, {\it potential} in GPI and {\it effective} in PI. The most promising split variables can be derived from predictors with a GPI greater than a fixed threshold. 
A fast algorithm can be adopted iterating the predictor selection with split identification until $\tau_{Y|s_l,Z}(t) \geq \tau_{Y|X_{l+1},Z}(t)$, where $s_l$ is the current split and $X_{l+1}$ the subsequent predictor in the rankings at step 1. Indeed, no other predictor can generate a split with higher PI. The split $s(t)$ with the highest PI and the predictor $X(t)$ that has generated it with the upper bound GPI are associated with the node $\{t\}$.

\subsection{Interpretation aids}
\label{subsec:aids}

In tree growth, the model-based split rule provides latent budgets parameters that update the response class profile of each level of the control variable within the child nodes. This comes from the Bayes interpretation of the latent budget parameter estimates\footnote{Omit the indication to the $g$-th group. Bayes' theorem states $\beta(j|k) = \frac{\pi_{+j}\pi_{k|j}}{\sum_j \pi_{+j}\pi_{k|j}} = \frac{\pi_{+j}\pi_{k|j}}{\pi_{k+}}$ where $\pi_{+j} = \sum_k \pi_{kj}$ and $\pi_{k+} = \sum_j \pi_{kj}$.}. 

At node $\{t\}$ in the $g$th group, ${\hat \beta}_{j|k(g)}(t)$ can be understood as the posterior probability estimate to assign the $j$-th response class once the observation is assigned to the left child node $\{2t\}$ for $k = 1$ or to the right child node $\{2t+1\}$ for $k = 2$ , while the conditional proportion $p_{j|(g)}(t)$ is the prior probability estimate. The Bayes factors representing the ratio between the posterior and prior probability estimates yield the following \emph{Lift} measures at node $\{t\}$:

\begin{equation}
\label{eq:lift}
    Lift_{j|k(g)}(t) = \frac{{\hat \beta}_{j|k(g)}(t)}{p_{j|(g)}(t)}
\end{equation}

\noindent
for the left child node $\{2t\}$ when $k = 1$ and the right child node $\{2t+1\}$ when $k = 2$. 
Each response class, within each group, is better predicted in the child node where the \emph{Lift} measure is the highest.

Starting from the root node of the tree ${\cal T}$, where $p_{j|(g)}(1)$ is the prior probability estimate belonging to class $j$ in the $g$-th group, the posterior probability estimate is recursively updated and related through a chain of conditional probability estimates until the final posterior classification $p_{j|(g)}(h)$ at the terminal node. The \emph{Leaf Classification Ratio} (LCR) is defined as:

\begin{equation}
\label{eq:LCR}
    LCR_{j|(g)}(h) = \frac{p_{j|(g)}(h)}{p_{j|(g)}(1)}
\end{equation}

\noindent
for any leaf $h$ in the set $h \in \tilde{\cal T}$.
The (\ref{eq:LCR}) allows identifying which response classes of the $g$ group are {\it overrepresented} in each leaf with respect to the root node. This is particularly useful in dealing with an unbalanced response distribution, to derive the tree paths, which are explanatory of the minority response classes.

\subsection{Visual pruning and decision tree selection} 
\label{subsec:vis}

The SLBT library offers the opportunity to interactively visualize all the information about the nodes and the tree paths. The tree-structure is represented using a node-link diagram, where the edges of the tree can have different lengths, each proportional to the impurity reduction provided by the split. This allows visualizing which split is more informative in any tree path. Furthermore, the node is represented as a pie chart showing the response class profile, particularly useful for a multi-class response variable. A dendrogram-like output with a non-decreasing function of the gain in information associated with the level of the tree allows pruning the {\it weak branches} for decision tree selection.

For any node $\{t\}$ of the tree ${\cal T}$ it is possible to define a measure of the proportional reduction in impurity of $Y$ obtained by the path that yields that descendant node as:

\begin{equation}
\label{eq:nodeIR}
    V_Y(t) = \frac{i_Y(1) - i_Y(t)}{i_Y(1)}
\end{equation}

\noindent
ranging in $[0,1]$, where $i_Y(1)$ and $i_Y(t)$ are the impurity measures of the response variable $Y$ at the root $\{1\}$ and the node $\{t\}$, respectively.

An overall proportional reduction of the impurity  of $Y$ in the final tree ${\cal T}$ can be obtained as a weighted sum of the reductions of the impurity in the leaves:

\begin{equation}
\label{eq:treeIR}
    V_Y({\cal T}) = \sum_{h \in \tilde {\cal T}} V_Y(h) p(h)
\end{equation}

\noindent
The (\ref{eq:treeIR}) can be used to select the decision tree adopting a visual pruning with an honest size selection procedure that cuts off branches causing overfitting\footnote{A statistical testing procedure based on a test sample can also be added to identify the honest size tree which is statistically reliable \citep{cappelli2002}.} \citep{iorio2019}. Once the expanded tree has been generated, the decision tree can be visually selected based on the predictability improvement (or impurity reduction) with respect to the complexity of the tree. More formally, a threshold measuring the purity gain of a tree when switching from a larger tree to a smaller one in terms of leaves can be fixed as:

\begin{equation}
\label{eq:treeDelta}
    \Delta({\cal T}_{\nu}) = \|V_Y({\cal T}_{\nu}) - V_Y({\cal T}_{\nu-1})\| \leq \varepsilon
\end{equation}

\noindent
where $V_Y({\cal T}_{\nu})$ and $V_Y({\cal T}_{\nu-1})$ are the proportional reduction measures of the impurity of the tree ${\cal T}_{\nu}$ and ${\cal T}_{\nu-1}$, respectively, and $\varepsilon$ is the threshold, typically set at $0.01$.

\subsection{Model choice} 
\label{subsec:modchoice}

SLBT adopts a recursive partitioning procedure where the split rule is based on a simultaneous mixture model, the SLBM in the unconstrained or constrained versions. The following four options are possible.

\begin{itemize}
    \item \textbf{SLBT-U:} the split rule considers the unconstrained model in which the sets of model parameters in the $G$ layers — both the mixing parameters and the latent budgets — are allowed to vary from layer to layer. This is equivalent to fitting a separate latent budget model with $K = 2$ latent components independently to each group of $Z$. In this flexible formulation, the growth of the tree proceeds recursively, highlighting in each node how the predictor acts differently in each level of $Z$, both in terms of the split rule and response profiles.
    \item \textbf{SLBT-A:} the split rule adopts the model with homogeneity mixing parameters while allowing the latent budget parameters to vary in the $G$ layers. In this formulation, the split rule driving the observations to each child node is forced to be the same for each group while highlighting different response profiles for each group. 
    \item \textbf{SLBT-B:} the split rule adopts the model with homogeneous latent budget parameters while allowing the mixing parameters to vary in the $G$ layers. Making the latent budgets, the expression of the response profiles, homogeneous implies inductive reasoning. To guaranty the same posterior probability estimates of the response classes in the different groups, how the split rule changes in different groups is highlighted.
    \item \textbf{SLBT-AB:} the split rule assumes that both the mixing and latent budget parameters are homogeneous across the $G$ layers. Assuming the same latent structure for the different groups leads to a recursive partition in which the stratification variable acts as a compound with the split.  
\end{itemize}

\noindent
The SLBT library offers the option that no stratification factor is specified, which means that LBT can be a special case. It should be noted that adding the stratification variable to the set of predictors in LBT would provide an unconditional tree structure.

\section{Least Squares Estimation based on Neural Network formulation}
\label{sec:LS}

SLBM can be estimated using the EM algorithm for the maximum likelihood method \citep{siciliano1994} or using the alternating least-squares algorithm \citep{Tambrea1999}. In this paper, a least squares estimation procedure is introduced built on a generalization, for the simultaneous case, of the approach originally proposed by Siciliano and Mooijaart for the neural network approach to the Latent Budget Model \citep{siciliano2001, yang2021}. One key advantage of this approach is that it avoids the computational burden associated with matrix inversion, which can be particularly intensive. However, this benefit comes at the potential cost of requiring more iterations before convergence is achieved.

SLBM can be understood as a set of supervised neural network models, specifically a set of double layer perceptrons, with linear activation functions, where weights are interpreted as conditional probabilities. In this perspective, the simultaneous latent budget structure as in figure \ref{fig: SLBT Split} is defined in terms of indicator variables instead of the contingency table. 
Consider, for the $g$-th layer, the $N_g \times I$ matrix ${\bf X}_g$ containing the $N_g$ observations of an input $I$-dimensional vector which includes the indicator variables associated to the predictor categories, the $N_g \times J$ matrix ${\bf Y}_g$ containing the $N_g$ observations of an output $J$-dimensional vector which includes the indicator variables associated to the response classes, the $I \times K$ matrix ${\bf A}_g$ of weights (mixing parameters) before the hidden layer and the $K \times J$ matrix ${\bf B}'_g$ of weights (latent budgets) following the hidden layer. The neural network is built to provide a classification rule to predict ${\bf Y}_g$ from ${\bf X}_g$ for a new observation of an unknown response class in simultaneous latent budget analysis.

The Simultaneous Neural Latent Budget Network (SNLBN) consists of a set of $G$ double-layer perceptrons, where the output can be defined as 

\begin{equation}
    y_{njg} = \psi_1 (\sum_k \beta_{j|k(g)} \psi_2(\sum_i x_{ni(g)}\alpha_{k|i(g)}) + d_{jg}) + e_{njg}
\end{equation}

\noindent
for any observation $n_g \in {\cal N}_g$ of the $g$-th sample, where $\psi_1$ and $\psi_2$ are the transfer functions associated with the output layer and the hidden layer, respectively.

Different transfer functions can be tried. Relating SLBM and SNLBN consider the identity transfer function and the null constant term $d_{jg}$ so that the output can be defined as:

\begin{equation}
    y_{njg} = \sum_k (\sum_i x_{ni(g)} \alpha_{k|i(g)})\beta_{j|k(g)} + e_{njg}
\end{equation}

\noindent
where the weights are nonnegative and satisfy suitable constraints.

In matrix formulation, the SNLBN can be defined as

\begin{equation}
\label{eq:SNLBN}
    {\bf Y}_g = {\bf X}_g{\bf A}_g{\bf B}'_g  + {\bf E}_g
\end{equation}

\noindent
where ${\bf E}_g$ is the error term, and ${\bf A}_g{\bf 1}_k = {\bf 1}_I$ and ${\bf 1}_J'{\bf B}_g = {\bf 1}_K'$. 

Pre-multiplying (\ref{eq:SNLBN}) by ${({\bf X}'_g{\bf X}_g)}^{-1}{\bf X}'_g$ yields SLBM (\ref{eq:SLBM})\footnote{Indeed, the SLBM in matrix formulation is given by $\bf \Pi = \bf AB'$, where $\bf \Pi$ is a $GI \times GJ$ block diagonal matrix composed of the theoretical budget matrices ${\bf \Pi}_g$, similarly the $GI \times GK$ block diagonal matrix $\bf A$ and the $GJ \times GK$ $\bf B$ with ${\bf A}_g$ and ${\bf B}_g$, respectively.}.

For the unconstrained model, for each $g$, both ${\bf A}_g$ and ${\bf B}_g$ are not constant. Thus, the function to optimize is:

\begin{equation}
\label{loss}
    f=\sum_g SSQ({\bf Y}_g - {\bf X}_g {\bf A}_g {\bf B}'_g) = \sum_g tr({\bf Y}'_g{\bf Y}_g - 2 {\bf Y}'_g {\bf X}_g {\bf A}_g {\bf B}'_g + {\bf B}_g {\bf A}'_g {\bf X}'_g {\bf X}_g {\bf A}_g {\bf B}'_g)
\end{equation}

\noindent
In (\ref{loss}), ${\bf X}'_g{\bf Y}_g$ is the frequency table ${\bf F}_g$; $tr({\bf Y}'_g{\bf Y}_g)= N_g$, $\sum_g Ng = N$; ${\bf X}'_g{\bf X}_g$ can be written as a diagonal matrix ${\bf D}_{Ig}$ with diagonal equal to ${\bf F}_{g}{\bf 1}_J$. The objective function (\ref{loss}) can be re-written as:

\begin{equation}
\label{eq:loss1}
    f = \sum_g SSQ({\bf Y}_g - {\bf X}_g {\bf A}_g {\bf B}'_g) 
    = N + \sum_g(-2tr({\bf F}'_{g}{\bf A}_{g}{\bf B}'_{g})+tr({\bf B}_{g}{\bf A}_{g}^{'}{\bf D}_{Ig}{\bf A}_{g}{\bf B}_{g}^{'}))
\end{equation}

When computing the derivatives of ${\bf F}_g$ with respect to ${\bf A}_g$ and ${\bf B}_g$, it should be noted that, for each $g$, only the components of the objective function associated with table $g$ depend on ${\bf A}_g$ and ${\bf B}_g$. Therefore, the partial derivatives can be computed independently for each $g$, as follows:

\begin{equation} 
    \partial f/\partial {\bf A}_g = -2{\bf F}_g{\bf B}_g + 2{\bf D}_{Ig}{\bf A}_g{\bf B}'_g{\bf B}_g 
\end{equation}
\begin{equation}
    \partial f/\partial {\bf B}_g = -2{\bf F}'_g{\bf A}_g + 2{\bf B}_g{\bf A}'_g {\bf D}_{Ig}{\bf A}_g
\end{equation}

\noindent
In order to have $\alpha_{k|i(g)} \geq 0$ with $\sum_k \alpha_{k|i(g)} = 1$ and $\beta_{j|k(g)} \geq 0$ with $\sum_j \beta_{j|k(g)} = 1$, the temporary matrices ${\bf U}_g$ and ${\bf V}_g$ can be used. The transformation that allows the transition from $\alpha_{k|i(g)}$ and $\beta_{j|k(g)}$ to $u_{ik(g)}$ and $v_{jk(g)}$, respectively, can be written as follows:

\begin{equation}
\label{eq:alfa}
    \alpha_{k|i(g)} = \exp{(u_{ik(g)})}/\sum_{m}\exp{(u_{im(g)}})
\end{equation}

\begin{equation}
\label{eq:beta}
    \beta_{j|k(g)} = \exp{(v_{jk(g)})}/\sum_{m}\exp{(v_{mk(g)}})
\end{equation}

Finally, the derivatives of $f$ with respect to ${\bf U}_g$ and ${\bf V}_g$ can be obtained by considering the following expressions:

\begin{equation}
    \partial f/\partial u_{ik(g)} = \sum_{l}(\partial f/\partial \alpha_{l|i(g)})(\delta^{lk}\alpha_{l|i(g)}-\alpha_{l|i(g)}\alpha_{k|i(g)}), k>l
\end{equation}

\begin{equation}
    \partial f/\partial v_{jk(g)} = \sum_{l}(\partial f/\partial \beta_{l|k(g)})(\delta^{lj}\beta_{l|k(g)}-\beta_{l|k(g)}\beta_{j|k(g)}), j>l
\end{equation}

\noindent
where $\delta$ is the usual Kronecker delta.

In constrained models, the loss function with the partial derivatives can be derived in straightforward way. The transition from constrained parameters to the values of the temporary matrices as in (\ref{eq:alfa}) and (\ref{eq:beta}) can be easily updated depending on the set of constraints.

The following procedure, applied with the appropriate equations depending on the model, can be used for all four forms of the SLBM. It consists of the following steps:

\begin{enumerate}
    \item Initialize the values for ${\bf U}_g$ and ${\bf V}_g$ as randomly generated;
    \item ${\bf U}_g$ is transformed to ${\bf A}_g$ and ${\bf V}_g$ is transformed to ${\bf B}_g$
    \item The derivatives of the objective function $f$ with respect to ${\bf A}_g$ and ${\bf B}_g$ are calculated;
    \item The gradient of $f$ with respect to ${\bf U}_g$ and ${\bf V}_g$ is computed and then ${\bf U}_g$ and ${\bf V}_g$ are updated;
    \item Iterate steps 2 to 4 until convergence is reached;
    \item After having solved ${\bf U}_g$ and ${\bf V}_g$ for $f$, the two matrices are transformed to ${\bf A}_g$ and ${\bf B}_g$.
\end{enumerate}

\section{ALS Case Study}
\label{sec:ALS}

Amyotrophic Lateral Sclerosis (ALS) is a rare progressive neurodegenerative disease defined by the selective attrition of motor neurons within the primary motor cortex, brain stem, and spinal cord. This neuronal degeneration involves a relentless decline in motor function that ultimately leads to death, usually due to respiratory failure \citep{wijesekera2009}. Although there is no cure, current treatments aim to prolong survival and quality of life. The life expectancy of a patient diagnosed with ALS is on average 3–5 years \citep{morris2015}. A recent systematic review \citep{wolfson2023} reported that ALS exhibits significant variability in prevalence and incidence at the global level, posing a major challenge for both healthcare systems and clinical modeling.
From a clinical perspective, the disease presents as evolving muscle weakness, which typically begins in the limb muscles (spinal onset), more frequently affecting distal rather than proximal muscles. Adverse prognostic factors include older age at onset, a faster disease progression rate, advanced clinical staging, and a reduced diagnostic delay, as earlier diagnosis often correlates with a more aggressive disease course. The high variability in the rate of progression and clinical phenotype makes it difficult to determine the optimal timing for medical interventions, such as non-invasive ventilation, and to ensure that a specific treatment is actually slowing the progression of the disease.

\subsection{Data understanding and exploration}

In this research, the SLBT models were applied exploring different homogeneity configurations as described in Section \ref{sec:SLBT} to the data set released by the ALS Center of the Federico II University Hospital of Naples, known as the \emph{ALS data set}. The cohort comprises $N=1412$ observations collected from $254$ patients, with baseline characteristics summarized in Table \ref{tab:raw_data}. This data set was previously investigated for adapting the Three-Tree Mixed-Effects model \citep{gottard2023} to longitudinal data, specifically 
to take into account temporal dependencies in the ALSFRS-R functional score \citep{vannucci2025}. The current analysis treats each clinical visit as an independent observation. The King's clinical stage was defined as the multi-class target variable $Y$, representing the sequential stages of disease spread and functional decline, while the patient's sex was selected as the stratification variable $Z \in \{Male,Female\}$. A comprehensive description of the variables considered in the analysis is reported in Table \ref{tab:variables_categorization}. All explanatory variables were processed in categorical format: original categorical or binary variables are reported with their natural modalities, while continuous variables were transformed into categorical variables using a K-means clustering algorithm based on the Elbow method (see details in Section \ref{app:lib}).

\begin{table}[h!]
\centering
\caption{Baseline and Clinical Characteristics of the ALS Dataset}
\label{tab:raw_data}
\begin{tabular}{llcc}
\toprule
\textbf{Variable} & \textbf{Category } & \textbf{N} & \textbf{\%} \\ 
\midrule
Total Observations & & 1412 & 100.0\% \\
\addlinespace
Gender & Male & 850 & 60.2\% \\
       & Female & 562 & 39.8\% \\
\addlinespace
Clinical Onset & Spinal & 1113 & 78.8\% \\
               & Bulbar & 269 & 19.0\% \\
               & Respiratory & 21 & 1.5\% \\
               & Multiple & 9 & 0.7\% \\
\addlinespace
King's Stage & Stage 0 & 6 & 0.4\% \\
             & Stage 1 & 244 & 17.3\% \\
             & Stage 2 & 310 & 22.0\% \\
             & Stage 3 & 414 & 29.3\% \\
             & Stage 4A & 20 & 1.4\% \\
             & Stage 4B & 418 & 29.6\% \\
\midrule \midrule
 & & \textbf{Mean} & \textbf{SD} \\ 
\midrule
Age at Onset (years) & & 58.06 & 12.58 \\
Diagnostic Delay (months) & & 17.80 & 17.72 \\
Disease Duration (months) & & 38.85 & 33.35 \\
\bottomrule
\end{tabular}
\end{table}

\begin{table}[h!]
	\centering
	\small
	\caption{Description of variables included in the analysis, divided into demographic, clinical, and disease progression domains. Variables highlighted in \textcolor{blue}{blue} were transformed into categorical variables using K-Means algorithm based on the Elbow method. For these variables, the Categories column reports the calculated centroids representing the representative value for each cluster, followed by the respective category modality in parentheses: Centroid (Modality). Original categorical or binary variables are reported with their natural modalities.}
	\label{tab:variables_categorization}
	\begin{tabular}{lp{4.5cm}l}
		\toprule
		\textbf{Variable} & \textbf{Description} & \textbf{Categories} \\ 
		\midrule
		\textit{Demographic} & & \\
		Sex & Patient's gender & Male (0), Female (1) \\
		\textcolor{blue}{Age at onset} & Age at symptom onset (years) & 34.1 (0), 49.6 (1), 60.4 (2), 72.8 (3) \\
		Family history & Family history of ALS & No (0), Yes (1) \\
		\midrule
		\textit{Clinical} & & \\
		Clinical onset & Initial manifestation type & Bulbar, Multiple, Respiratory, Spinal \\
		FVC & Forced Vital Capacity (\%) & $<$40, 40-60, 60-80, $>$80, ND \\
		VENTILATION & Use of non-invasive ventilation & 0: None, 1: Active, 2: Other pathol., \\
		             &                                & 3: Indicated but not performed \\
		Tracheostomy & Presence of tracheostomy & No (0), Yes (1) \\
		PEG & Presence of Gastrostomy & 0 (No), 1 (Yes), 2 (Refused) \\
		\textcolor{blue}{MRC UL} & Upper Limb strength (0-70) & 4.8 (0), 34.8 (1), 62.3 (2) \\
		\textcolor{blue}{MRC LL} & Lower Limb strength (0-60) & 5.3 (0), 32.3 (1), 54.5 (2) \\
		\textcolor{blue}{MRC Bulbar} & Bulbar strength (0-15) & 0.3 (0), 7.1 (1), 11.9 (2), 14.6 (3) \\
		\textcolor{blue}{PUMNS UL} & UMN burden Upper Limbs & 0.5 (0), 4.6 (1), 9.2 (2) \\
		\textcolor{blue}{PUMNS LL} & UMN burden Lower Limbs & 0.6 (0), 3.9 (1), 7.2 (2), 11.5 (3) \\
		\textcolor{blue}{PUMNS Bulbar} & UMN burden Bulbar region & 0 (0), 1.4 (1), 3 (2), 4 (3) \\
		\textcolor{blue}{CNS LS} & CNS involvement score & 7.1 (0), 16.9 (1), 30.1 (2) \\
		Lability Scale & Cut off lability scale & No (0), Yes (1) \\
		\midrule
		\textit{Progression} & & \\
		\textcolor{blue}{Diagnostic delay} & Time onset to diagnosis (months) & 8.8 (0), 25.0 (1), 57.0 (2), 96.6 (3) \\
		\textcolor{blue}{Disease duration} & Duration since onset (months) & 17.1 (0), 43.7 (1), 84.4 (2), 144.2 (3) \\
		\textcolor{blue}{Progression rate} & Speed of progression & 0.4 (0), 1.4 (1), 3.5 (2), 7.9 (3) \\
		Therapy & Ongoing ALS-specific therapy & Riluzolo, Edaravone, Both, None \\
        		\midrule
            {King's Stage} & {Target Variable} & {0, 1, 2, 3, 4A, 4B} \\
	\bottomrule
	\end{tabular}
\end{table}

\subsection{SLBT Results}

To investigate the complex association between clinical predictors and ALS progression, four configurations of the SLBT algorithm are compared, each reflecting different assumptions regarding the homogeneity of disease progression between genders:

\begin{itemize}
	\item \textbf{SLBT-U:} This model allows for total heterogeneity, with both mixing parameters (split rules) and latent budget parameters (response profiles) estimated independently for each group.
	\item \textbf{SLBT-A:} This configuration enforces a common tree structure with the same split rules across groups while allowing the latent budget parameters $\beta_{j|k(g)}(t)$ to vary.
	\item \textbf{SLBT-B:} In this case, the tree structure is allowed to diverge between groups but enforces the same latent budget parameters, testing if gender-specific predictors lead to the same outcomes.
	\item \textbf{SLBT-AB:} This model assumes that both the split rules and the latent budget structure are identical across the $G$ layers.
\end{itemize}

\begin{figure}[ht]
    \centering
    \includegraphics[width=1.0\linewidth]{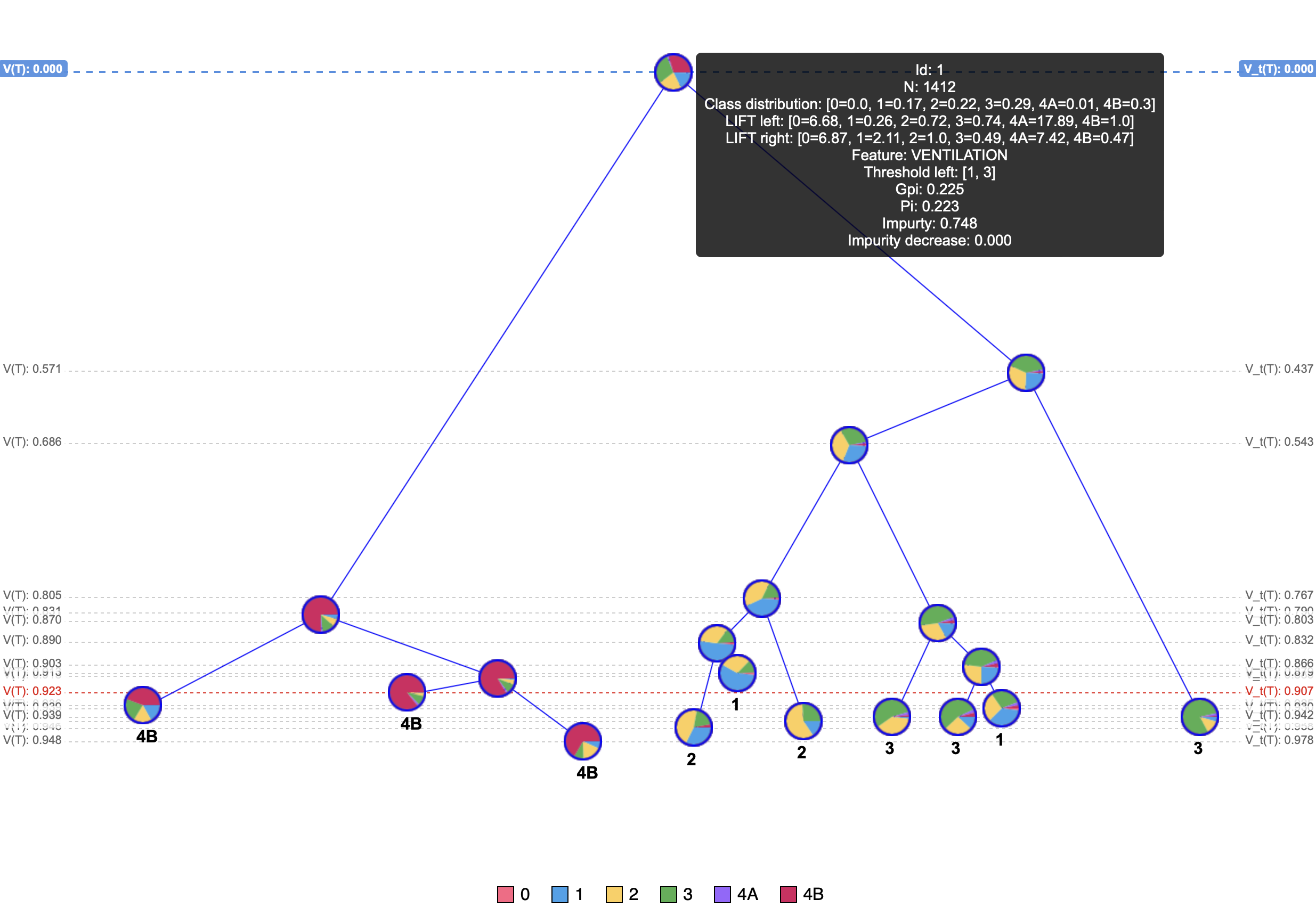}
    \caption{SLBT-AB pruned classification tree for King’s staging in ALS patients.}
    \label{fig:SLBT_AB}
\end{figure}

\begin{figure}[ht]
    \centering
    \includegraphics[width=1.0\linewidth]{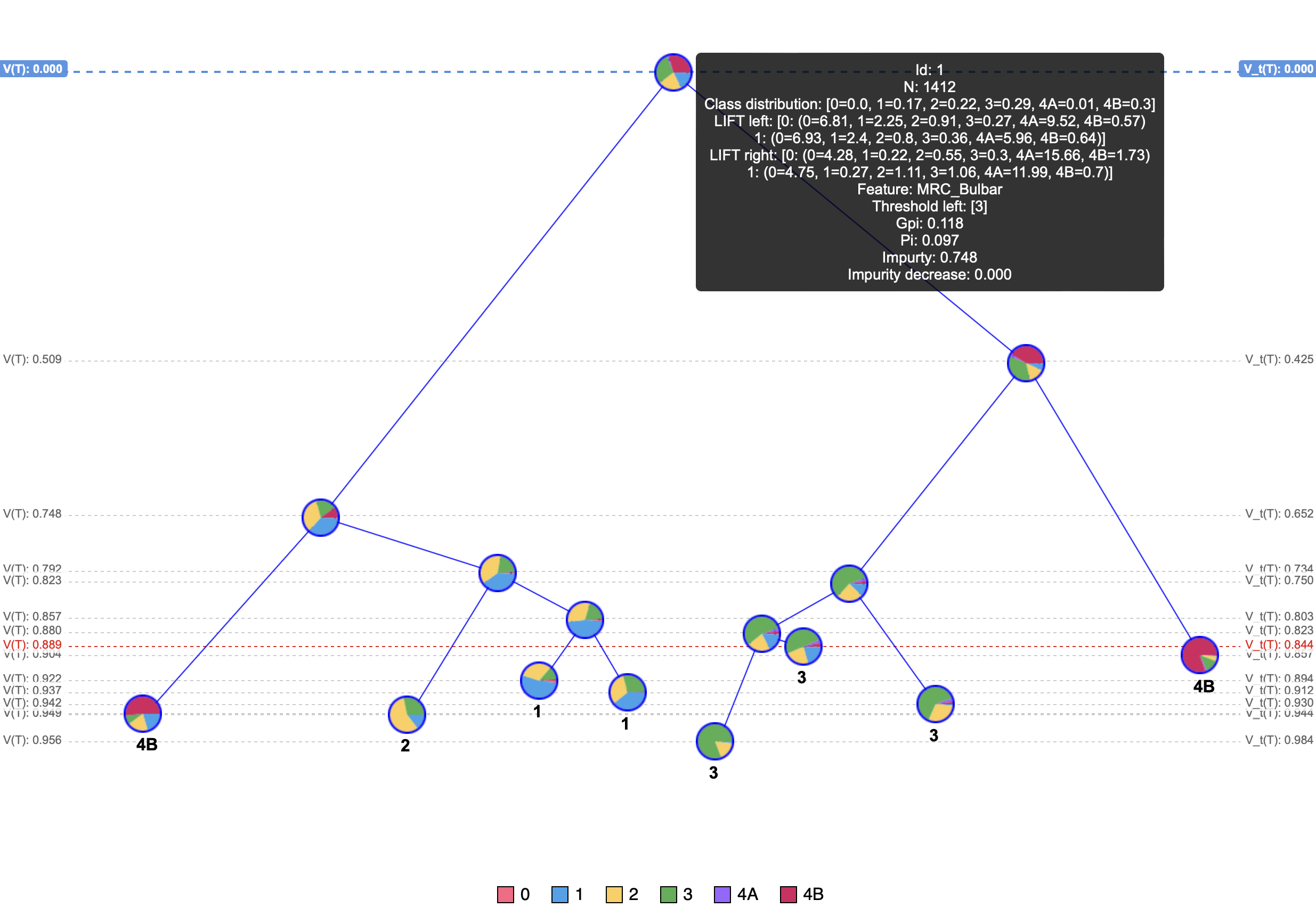}
    \caption{SLBT-A pruned classification tree for King’s staging in ALS patients.}
    \label{fig:SLBT_A}
\end{figure}

For clarity and to facilitate clinical interpretation, the tree structures of the pruned SLBT-AB model and the pruned SLBT-A model are presented in Figure \ref{fig:SLBT_AB} and \ref{fig:SLBT_A}, respectively. The complete set of tree structures and results for all four configurations, including SLBT-U and SLBT-B as well as all versions of maximum expanded trees before pruning, is provided for interactive exploration in the dedicated GitHub repository at \url{https://github.com/Giugurtah/SLBT}.

From the SLBT-AB model (Figure \ref{fig:SLBT_AB}), it is possible to describe the high-severity scenario identified by the leftmost branches of the tree, for which the prediction at the terminal nodes is the late-stage of the disease. In the root node $\{1\}$, the prior probabilities for the most severe stages are $0.01$ for Stage $4A$ and $0.30$ for Stage $4B$. The model identifies VENTILATION as the splitting predictor. For a patient requiring non-invasive ventilation or ventilation indicated but not performed (modalities 1 and 3) there is a probability update for both stages. The discriminative power of the split is evident when comparing the two child nodes: \emph{Lift left} for stage $4A$ is $17.89$, which is more than twice as \emph{Lift right} ($7.41$) for patients who do not require ventilation. For stage $4B$, \emph{Lift left} is $1.00$, while \emph{Lift right} drops to $0.46$. This indicates that the absence of ventilation needs reduces the odds of being in stage $4B$ more than $50\%$ compared to the root node. Further down in this scenario, the model utilizes FVC and MRC Bulbar to refine the predictions. Among patients with VENTILATION $\in \{1,3\}$, those who maintain high values of FVC ($> 80\%$) go to the terminal node $\{4\}$ with $LCR$ for King's stages are $1= 0.97, 2= 0.77, 3=0.75, 4B= 1.49$. In contrast, patients with lower FVC values are further stratified through MRC Bulbar. Here, lower scores of MRC Bulbar define the cases at the terminal node $\{10\}$, where the $LCR$ in King's stages are $1= 0.03, 2= 0.14, 3=0.31, 4A=1.3, 4B= 2.89$, while a high score of MRC Bulbar defines the cases at the terminal node $\{11\}$, where the $LCR$ in King's stages are $1= 0.39, 2= 0.81, 3=0.30, 4B= 2.25$. Ultimately, the analysis of these scenarios highlights that while ventilation identifies high-risk patients, it is the cumulative impact of additional variables such as FVC and MRC bulbar that can help in defining the most severe critical profiles.  

Figure \ref{fig:SLBT_A} shows the pruned classification tree for the SLBT-A model, which investigates gender-specific patterns by adopting a heterogeneous configuration in terms of latent budgets. In the root node $\{1\}$, the impact of MRC Bulbar on the stage of the disease shows immediate gender differences. For patients with MRC Bulbar lower than $3$, the \emph{Lift right} towards Stage 4B is significantly higher for males ($1.73$) than for females ($0.70$). Following the extreme right branch of this tree, patients requiring non-invasive ventilation or ventilation indicated but not performed (modalities 1 and 3) lead directly to the terminal node $\{7\}$. Interestingly, the \emph{Lift} associated with this split is higher in females ($1.20$) compared to males (0.23) for stage $4B$. This suggests that for women, the coexistence of bulbar and respiratory impairment is an even more direct indicator of the terminal phase. Instead, for patients who do not yet require ventilation, the model uses the MRC Upper Limb to refine the predictions. In node $\{13\}$, a declining profile is observed, with high concentrations of stage 3 ($LCR = 2.14$) and stage 4A ($LCR = 2.43$). The \emph{Lift} measures at node $\{6\}$ show that MRC Upper Limb $\in \{0, 2\}$ has a stronger impact on females for stage 4B (\emph{Lift left} 10.50) than on males (\emph{Lift left} 7.84), indicating a different prioritization of motor symptoms between genders during the decline. Finally, for patients characterized by bulbar impairment and the absence of ventilatory support, the model introduces Disease Duration as a splitting predictor. This split further refines the LCR in the terminal nodes, revealing how the time factor interacts with functional loss. For patients with a longer duration of the disease (disease duration $\in \{2,3\}$) in the node $\{24\}$, there is a significant representation of stage 3 ($LCR = 2.75$ representing approximately 80\% of the node distribution). In contrast, in patients with a lower disease history (disease duration $\in \{0,1\}$) in node 25, even if stage 3 remains the most frequent (representing about 50\% of the node distribution), stage 1 and stage 2 together account for approximately 45\% of the distribution. Therefore, this terminal node identifies patients with lower values of MRC Upper Limb, and for whom the shorter duration of the disease makes it significantly more likely that they are still in the early stages of the disease compared to those with a longer history. In conclusion, analysis of the SLBT-A model demonstrates that although the fundamental biological stages of ALS are common, the clinical weight and timing of these transitions are highly sensitive to gender, allowing for a more personalized identification of the most critical clinical profiles.

\section{Concluding remarks}
\label{sec:con}

This paper introduced \emph{Simultaneous Latent Budget Trees} (SLBT) as an explainable probabilistic machine leaning model specifically designed to address classification problems in which a stratification factor acts as control variable.
The main strength of the SLBT methodology lies in its capacity to simultaneously evaluate the explanatory power of predictors across the multi-dimensional structure of a data cube. By embedding the Simultaneous Latent Budget Model directly into the recursive partitioning process as a model-based splitting criterion, the algorithm transcends the traditional limitations of standard CART. Indeed, by simultaneously modeling the data partition and the latent structure of the response, a transparent assessment of how predictors interact with the stratification factor is obtained. The introduction of least squares estimation based on a neural network formulation also ensures computational efficiency while maintaining high interpretability.  

The SLBT library has been developed that features the implementation of an interactive decision tree interface. This tool allows users to instantly access all statistical measures proposed for each specific node simply by hovering the cursor over the node of interest. Moreover, the library incorporates statistical indicators such as \emph{Global Predictability Improvement} and \emph{Predictability Improvement} which offer a deeper evaluation of information gain compared to standard metrics. This kind of measure is currently not present in mainstream tree-based libraries. In addition, the library includes an interactive visual pruning procedure, which provides a unified framework for model selection. By displaying information trees in which the length of the branches is proportional to the reduction in impurity, the most predictive paths can be immediately evaluated.

The methodology was applied to a set of real ALS data, considering King's Stage as a multi-class target variable, a set of demographic and clinical characteristics as explanatory variables, and gender as a stratification factor. The different model configurations proposed were compared, from the unconstrained version to those with homogeneity constraints on mixing parameters or latent budgets. The analysis suggests that considering gender as a stratification factor improves the ability of the model to capture the clinical heterogeneity of the duration of the ALS progression.

In particular, the constrained SLBT-A model showed that bulbar impairment serves as the most immediate indicator of terminal stages in males, while upper limb decline has a stronger impact on the progression trajectory of females. The use of \emph{Lift} and \emph{LCR} allows the identification of specific clinical profiles even in minority stages. It should be noted that the SLBT model is highly flexible and can be applied using different stratification variables in addition to gender. For example, the number of visits or the time to onset could be considered as a stratification factor. In this way, it would be possible to reveal how decision-making pathways evolve over the course of the disease. This would allow specific time windows to be identified in which variables acquire or lose predictive power, providing a more dynamic map of the progression of ALS.

Future developments will focus on the expansion of the SLBT to naturally handle numerical predictors. Since the SNLBN formulation is based on a neural network framework, adapting the input layer will allow for a unified treatment of both categorical and continuous data without preliminary categorization. Further research will also include the implementation of other 
non-standard tree growth procedures, ensemble methods, and a standard cost-complexity pruning procedure based on V-fold cross-validation. Finally, the methodology aims to be extended to longitudinal data cubes, enabling explicit modeling of temporal dependencies over time.

\appendix

\backmatter

\section*{Statements and Declarations}

\subsection*{Funding}

This work was supported by the Italian Ministry of Research, in complementary actions to the NRRP \virgolette{Fit4MedRob - Fit for Medical Robotics} Grant (\# PNC0000007). 

\subsection*{Data Availability Statement}
The ALS data set analyzed during the current study is not publicly available due to privacy and ethical restrictions related to patient data. However, the data are available from the corresponding author upon request. The complete source code, including the implementation of the SLBT algorithm and the scripts for tree pruning and visualization, is openly available in the GitHub repository at \url{https://github.com/Giugurtah/SLBT}.

\subsection*{Competing Interests}
The authors have no relevant financial or non-financial interests to disclose.

\subsection*{Acknowledgments}
The authors would like to thank the team of Prof. Raffaele Dubbioso at the ALS Center of Federico II University Hospital of Naples for providing the data set used in this study.

\bibliography{bibfile}

\begin{thebibliography}{45}
\providecommand{\natexlab}[1]{#1}
\providecommand{\doi}[1]{\url{https://doi.org/#1}}
\providecommand{\url}[1]{\texttt{#1}}
\providecommand{\urlprefix}{}

\bibitem[{Breiman(1996)Breiman, Leo}]{breiman1996}
Breiman L.
\newblock Bagging Predictors.
\newblock Machine Learning. 1996;24(2):123--140.

\bibitem[{Breiman(1998)Breiman, Leo}]{breiman1998}
Breiman L.
\newblock Arcing Classifiers.
\newblock The Annals of Statistics. 1998;26(3):801--849.

\bibitem[{Breiman(2001)Breiman, Leo}]{breiman2001}
Breiman L.
\newblock Random Forests.
\newblock Machine Learning. 2001;45(1):5--32.

\bibitem[{Breiman et~al.(1984)Breiman, Leo and Friedman, Jerome H. and Olshen,
  Richard A. and Stone, Charles J.}]{breiman1984}
Breiman L, Friedman JH, Olshen RA, Stone CJ.
\newblock Classification and Regression Trees.
\newblock New York: Chapman and Hall/CRC; 1984.

\bibitem[{Cappelli et~al.(2002)Cappelli, C. and Mola, F. and Siciliano,
  R.}]{cappelli2002}
Cappelli C, Mola F, Siciliano R.
\newblock A statistical approach to growing a reliable honest tree.
\newblock Computational Statistics \& Data Analysis. 2002;38(3):285--299.

\bibitem[{Conversano(2011)Conversano, Claudio}]{conversano2011}
Conversano C.
\newblock Interactive visualization in multiclass learning: integrating the
  SASSC algorithm with KLIMT.
\newblock Computational Statistics. 2011;26:711--731.

\bibitem[{Conversano and Dusseldorp(2017)Conversano, Claudio and Dusseldorp,
  Elise}]{conversano2017}
Conversano C, Dusseldorp E.
\newblock Modeling Threshold Interaction Effects through the Logistic
  Classification Trunk.
\newblock Journal of Classification. 2017;34(3):399--426.

\bibitem[{Cotugno et~al.(2025)Cotugno, Lorenza and Pellegrino, Stefano and
  Siciliano, Roberta}]{cotugno2025}
Cotugno L, Pellegrino S, Siciliano R.
\newblock Decoding Locomotor Intentions: Neural Networks and Probabilistic
  Machine Learning for Customizable Exoskeletons.
\newblock In: {D'Ambrosio} A, de~Rooij M, {De Roover} K, Iorio C, {La Rocca} M,
  editors. Supervised and Unsupervised Statistical Data Analysis Studies in
  Classification, Data Analysis, and Knowledge Organization, Cham: Springer;
  2025. p. 107--118.

\bibitem[{{D'Ambrosio} and Heiser(2019){D'Ambrosio}, Antonio and Heiser, Willem
  J.}]{dambrosio2016}
{D'Ambrosio} A, Heiser WJ.
\newblock A recursive partitioning method for the prediction of preference
  rankings based upon Kemeny distances.
\newblock Psychometrika. 2019;81(3):774--794.

\bibitem[{Dusseldorp et~al.(2010)Dusseldorp, Elise and Conversano, Claudio and
  Van Os, Bart Jan}]{dusseldorp2010}
Dusseldorp E, Conversano C, Van~Os BJ.
\newblock Combining an Additive and Tree-Based Regression Model Simultaneously:
  STIMA.
\newblock Journal of Computational and Graphical Statistics.
  2010;19(3):514--530.

\bibitem[{Dusseldorp and Meulman(2004)Dusseldorp, Elise and Meulman, Jacqueline
  J.}]{dusseldorp2004}
Dusseldorp E, Meulman JJ.
\newblock The Regression Trunk Approach to Discover Treatment Covariate
  Interaction.
\newblock Psychometrika. 2004;69(3):355--374.

\bibitem[{Fokkema et~al.(2021)Fokkema, Marjolein and Edbrooke-Childs, Julian
  and Wolpert, Miranda}]{fokkema2021}
Fokkema M, Edbrooke-Childs J, Wolpert M.
\newblock Generalized Linear Mixed-Model (GLMM) Trees: A Flexible Decision-Tree
  Method for Multilevel and Longitudinal Data.
\newblock Psychotherapy Research. 2021;31(3):329--341.

\bibitem[{Friedman et~al.(2000)Friedman, Jerome and Hastie, Trevor and
  Tibshirani, Robert and others}]{friedman2000}
Friedman J, Hastie T, Tibshirani R, et~al.
\newblock Additive Logistic Regression: a Statistical View of Boosting (with
  Discussion and a Rejoinder by the Authors).
\newblock The Annals of Statistics. 2000;28(2):337--407.

\bibitem[{Goodman and Kruskal(1954)Goodman, L. A. and Kruskal, W.
  H.}]{goodman1954}
Goodman LA, Kruskal WH.
\newblock Measures of association for cross classification.
\newblock Journal of the American Statistical Association. 1954;48(2):732--762.

\bibitem[{Gottard et~al.(2023)Gottard, Anna and Vannucci, Giulia and Grilli,
  Leonardo and Rampichini, Carla}]{gottard2023}
Gottard A, Vannucci G, Grilli L, Rampichini C.
\newblock Mixed-Effect Models with Trees.
\newblock Advances in Data Analysis and Classification. 2023;17(2):431--461.

\bibitem[{Gray and Williams(1975)Gray, L. N. and Williams, J. S.}]{gray1975}
Gray LN, Williams JS.
\newblock Goodman and Kruskal's Tau b: Multiple and Partial Analogs.
\newblock In Proceedings of the Social Statistics Section, Journal of the
  American Statistical Association. 1975;p. 444--448.

\bibitem[{Grubinger et~al.(2014)Grubinger, Thomas and Zeileis, Achim and
  Pfeiffer, Karl-Peter}]{grubinger2014}
Grubinger T, Zeileis A, Pfeiffer KP.
\newblock evtree: Evolutionary Learning of Globally Optimal
  Classification and Regression Trees in R.
\newblock Journal of Statistical Software. 2014;61:1--29.

\bibitem[{Hastie et~al.(2009)Hastie, Trevor and Tibshirani, Robert and
  Friedman, Jerome}]{hastie2009}
Hastie T, Tibshirani R, Friedman J.
\newblock The Elements of Statistical Learning: Data Mining, Inference, and
  Prediction, vol.~2.
\newblock New York: Springer; 2009.

\bibitem[{Hothorn et~al.(2006)Hothorn, Torsten and Hornik, Kurt and Zeileis,
  Achim}]{hothorn2006}
Hothorn T, Hornik K, Zeileis A.
\newblock Unbiased Recursive Partitioning: A Conditional Inference Framework.
\newblock Journal of Computational and Graphical Statistics.
  2006;15(3):651--674.

\bibitem[{Iorio et~al.(2019)Iorio, Carmela and Aria, Massimo and {D'Ambrosio},
  Antonio and Siciliano, Roberta}]{iorio2019}
Iorio C, Aria M, {D'Ambrosio} A, Siciliano R.
\newblock Informative Trees by Visual Pruning.
\newblock Expert Systems with Applications. 2019;127:228--240.

\bibitem[{Jelihovschi and Allaman(2018)Jelihovschi, Enio G. and Allaman, Ivan
  Bezerra}]{Jelihovschi2018}
Jelihovschi EG, Allaman IB.
\newblock lba: An R Package for Latent Budget Analysis.
\newblock The R Journal. 2018;10(1):269--287.

\bibitem[{Kim and Kim(2022)Kim, Ahhyoun and Kim, Hyunjoong}]{kim2022}
Kim A, Kim H.
\newblock A New Classification Tree Method With Interaction Detection
  Capability.
\newblock Computational Statistics \& Data Analysis. 2022;165:107324.

\bibitem[{Kindo et~al.(2016)Kindo, Bereket P. and Wang, Hao and Pe{\~n}a, Edsel
  A.}]{kindo2016}
Kindo BP, Wang H, Pe{\~n}a EA.
\newblock Multinomial Probit Bayesian Additive Regression Trees.
\newblock Stat. 2016;5(1):119--131.

\bibitem[{de~Leeuw and {van der}~Heijden(1988)de Leeuw, Jan and {van der}
  Heijden, Peter G. M.}]{deleeuw1988}
de~Leeuw J, {van der}~Heijden PGM.
\newblock The Analysis of Time-Budgets with a Latent Time-Budget Model.
\newblock Data Analysis and Informatics 5 (ed E Diday). 1988;p. 159--166.

\bibitem[{Light and Margolin(1971)Light, Richard J. and Margolin, Barry
  H.}]{light1971}
Light RJ, Margolin BH.
\newblock An Analysis of Variance for Categorical Data.
\newblock Journal of the American Statistical Association. 1971;66:534--544.

\bibitem[{Loh and Zhou(2020)Loh, Wei-Yin and Zhou, Peigen}]{loh2020}
Loh WY, Zhou P.
\newblock In: Ting N, Cappelleri JC, Ho S, Chen DDG, editors. The GUIDE
  Approach to Subgroup Identification Cham: Springer International Publishing;
  2020. p. 147--165.

\bibitem[{Mola and Siciliano(1992)Mola, Francesco and Siciliano,
  Roberta}]{mola1992}
Mola F, Siciliano R.
\newblock A Two-Stage Predictive Splitting Algorithm in Binary Segmentation.
\newblock In: Computational Statistics: Volume 1: Proceedings of the 10th
  Symposium on Computational Statistics Springer; 1992. p. 179--184.

\bibitem[{Mola and Siciliano(1997)Mola, Francesco and Siciliano,
  Roberta}]{mola1997}
Mola F, Siciliano R.
\newblock A Fast Splitting Procedure for Classification Trees.
\newblock Statistics and Computing. 1997;7:209--216.

\bibitem[{Mooijaart et~al.(1999)Mooijaart, A. and {van der} Heijden, P. G. M.
  and {van der Ark}, L. A.}]{mooijaart1999}
Mooijaart A, {van der}~Heijden PGM, {van der Ark} LA.
\newblock A least squares algorithm for a mixture model for compositional data.
\newblock Computational Statistics and Data Analysis. 1999;30:359--379.

\bibitem[{Morris(2015)Morris, Jerry}]{morris2015}
Morris J.
\newblock Amyotrophic lateral sclerosis (ALS) and related motor neuron
  diseases: an overview.
\newblock The Neurodiagnostic Journal. 2015;55(3):180--194.

\bibitem[{Seibold et~al.(2016)Seibold, Heidi and Zeileis, Achim and Hothorn,
  Torsten}]{seibold2016}
Seibold H, Zeileis A, Hothorn T.
\newblock Model-based recursive partitioning for subgroup analyses.
\newblock The International Journal of Biostatistics. 2016;12(1):45--63.

\bibitem[{Siciliano(1999)Siciliano, Roberta}]{siciliano1999}
Siciliano R.
\newblock Latent Budget Trees for Multiple Classification.
\newblock In: Classification and Data Analysis: Theory and Application.
  Proceedings of the Biannual Meeting of the Classification Group of
  Societ{\`a} Italiana di Statistica (SIS), Pescara, July 3--4, 1997 Springer;
  1999. p. 121--128.

\bibitem[{Siciliano and Mola(2000)Siciliano, Roberta and Mola,
  Francesco}]{siciliano2000}
Siciliano R, Mola F.
\newblock Multivariate Data Analysis and Modeling through Classification and
  Regression Trees.
\newblock Computational Statistics \& Data Analysis. 2000;32(3-4):285--301.

\bibitem[{Siciliano and Mooijaart(2001)Siciliano, Roberta and Mooijaart,
  Ab}]{siciliano2001}
Siciliano R, Mooijaart A.
\newblock Unconditional Latent Budget Analysis: a Neural Network Approach.
\newblock In: Advances in Classification and Data Analysis Springer; 2001. p.
  127--134.

\bibitem[{Siciliano and {van der}~Heijden(1994)Siciliano, Roberta and {van der}
  Heijden, Peter G. M.}]{siciliano1994}
Siciliano R, {van der}~Heijden PGM.
\newblock Simultaneous latent budget analysis of a set of two-way tables with
  constant-row-sum data.
\newblock Metron. 1994;53:155--179.

\bibitem[{Tambrea and Siciliano(1999)Tambrea, Narcisa and Siciliano,
  Roberta}]{Tambrea1999}
Tambrea N, Siciliano R.
\newblock Exploratory Analysis of Three-way Data by Simultaneous Latent Budget
  Model.
\newblock Applied Stochastic Models in Business and Industry. 1999;15:469--484.

\bibitem[{{van der}~Heijden et~al.(1989){van der} Heijden, Peter G. M. and
  Mooijaart, Ab and de Leeuw, Jan}]{vanderHeijden1989}
{van der}~Heijden PGM, Mooijaart A, de~Leeuw J.
\newblock Latent Budget Analysis.
\newblock In: Decarli, A, Francis, B J, Gilchrist, R, Seeber, G U H (eds):
  Statistical Modelling Lecture Notes in Statistics. 1989;57:301--313.

\bibitem[{Vannucci and Gottard(2023)Vannucci, Giulia and Gottard,
  Anna}]{vannucci2023}
Vannucci G, Gottard A.
\newblock An Evolutionary Estimation Procedure for Generalized Semilinear
  Regression Trees.
\newblock Computational Statistics. 2023;38(4):1927--1946.

\bibitem[{Vannucci et~al.(2025)Vannucci, Giulia and Siciliano, Roberta and
  Iuzzolino, Valentina and Senerchia, Gianmaria and Dubbioso,
  Raffaele}]{vannucci2025}
Vannucci G, Siciliano R, Iuzzolino V, Senerchia G, Dubbioso R.
\newblock Enhancing Statistical Inference in Mixed-Effect Three-Tree Model: A
  Data-Carving Estimation Strategy with an Application on Amyotrophic Lateral
  Sclerosis Data.
\newblock In: {D'Ambrosio} A, de~Rooij M, {De Roover} K, Iorio C, {La Rocca} M,
  editors. Supervised and Unsupervised Statistical Data Analysis Cham: Springer
  Nature Switzerland; 2025. p. 341--352.

\bibitem[{Vapnik(1995)Vapnik, Vladimir N.}]{vapnik1995}
Vapnik VN.
\newblock The Nature of Statistical Learning Theory.
\newblock New York: Springer; 1995.

\bibitem[{Vapnik(1998)Vapnik, Vladimir N.}]{vapnik1998}
Vapnik VN.
\newblock Statistical Learning Theory.
\newblock New York: Wiley; 1998.

\bibitem[{Wijesekera and Leigh(2009)Wijesekera, Lokesh C. and Leigh, P.
  Nigel}]{wijesekera2009}
Wijesekera LC, Leigh PN.
\newblock Amyotrophic lateral sclerosis.
\newblock Orphanet Journal of Rare Diseases. 2009;4(1):3.

\bibitem[{Wolfson et~al.(2023)Wolfson, Christina and Gauvin, Danielle E. and
  Ishola, Foluso and Oskoui, Maryam}]{wolfson2023}
Wolfson C, Gauvin DE, Ishola F, Oskoui M.
\newblock Global prevalence and incidence of amyotrophic lateral sclerosis: a
  systematic review.
\newblock Neurology. 2023;101(6):e613--e623.

\bibitem[{Yang et~al.(2021)Zhenwei Yang and Ayoub Bagheri and P. G. M van der
  Heijden}]{yang2021}
Yang Z, Bagheri A, van~der Heijden PGM.: Neural Networks for Latent Budget
  Analysis of Compositional Data; 2021.
\newblock \urlprefix\url{https://arxiv.org/abs/2109.04875}.

\bibitem[{Zhou(2012)Zhou, Zhi-Hua}]{zhou2012}
Zhou ZH.
\newblock Ensemble Methods: Foundations and Algorithms.
\newblock New York: CRC Press; 2012.

\end{thebibliography}

\section{The SLBT library}
\label{app:lib}

\subsection{The SLBT class}
The SLBT class is the principal component of the library, it implements both the standard Simultaneous Latent Budget Tree and the Latent Budget Tree as a particular case of the former based on the presence of a stratification variable.

\subsubsection{Constructor Parameters}
The class can be initialized with the following parameters:
\begin{itemize}
    \item \textit{homogeneity}: it controls the homogeneity constraints that are applied during the execution of the model. The allowed values are "none", "A", "B" and "AB". They force the unconstrained model, the across-group homogeneity constraints for the mixing parameters, the across-group homogeneity constraints for the latent budget parameters, and the across-group homogeneity constraints for the latent budget structure, respectively. The default, if not given, is "none".
    \item \textit{max\_depth}: it controls the maximal depth of the tree. During the fitting process when a branch of the tree reaches the maximal depth, the last node is set as a leaf node. The default, if not given, is 10.
    \item \textit{min\_pi}: it controls the minimal Predictability Index. If the best splitting variable is associated with a pi lower than the set minimum, the splitting is not performed, and the node is set as a leaf node. The default, if not given, is 0.
    \item \textit{min\_gpi}: it controls the minimal Global Predictability Index. Before the splitting process, if a node has a gpi lower than the set minimum, the splitting process is not performed and the current node is set as a leaf node. The default, if not given, is 0.
    \item \textit{min\_impurity}: it controls the minimal impurity. Before the splitting process, if the current node has an impurity lower than the set minimum, the splitting process is not performed, and the current node is set as a leaf node. This ensures an early stop on the nearly pure nodes. The default, if not given, is 0.
    \item \textit{min\_samples\_split}: it controls the minimal number of data points that a node has to contain to split it. Before the splitting process, if the current node contains a lower number of data points than the set minimum, the splitting process is not performed, and the current node is set as a leaf node. The default, if not given, is 2.
\end{itemize}
Setting all constructor parameters, except for homogeneity, at their default values ensures a fully extended tree. 

\subsubsection{Fitting Method}
Once the class has been initialized, the SLBT model is fit to the training data with the "fit(X, y, x\_s)" method.
\begin{itemize}
    \item \textit{X}: the training features given as a pandas.DataFrame of shape $N$ samples and $K$ features. All features must be categorical of any type. For continuous variables, use the Categorizer class for pre-processing.
    \item \textit{y}: the target variable given as a pandas.Series of shape N samples. The variable must be categorical of any type.
    \item \textit{x\_s}: the stratification variable given as a pandas.Series of shape N samples. The variable must be categorical of any type. This variable is optional and, when provided, enables the SLBT model with the specified homogeneity constraint. If not provided, it enables the LBT model.
\end{itemize}

\subsubsection{Interactive Visualization}
The library also provides an HTML-based interactive visualization interface for exploring tree-structures and analyzing decision rules. This can be achieved with the "plot\_html(model, output\_file, title, color\_palette, visual\_pruning)" method. 
\begin{itemize}
    \item \textit{model}: the SLBT fitted model to visualize.
    \item \textit{output\_file}: the output HTML file path. It must be a string. If not provided, the default is "tree\_visualization.html"
    \item \textit{title}: the title to print on the final visualization interface. Must be a string. If not provided, the default is "Decision Tree Visualization".
    \item \textit{color\_palette}: optional list of hex colors for class visualization. It must be a list of strings. If the number of modalities of the target variable is less than the number of colors provided, only the first required colors are used. If the number of modalities is higher, some classes will share the same color. If not provided, the default uses a predetermined set of nine colors.
    \item \textit{visual\_pruning}: it enables visual pruning visualization. It must be a boolean (True/False) input. If not provided, the default is True.
\end{itemize}

\subsection{The Categorizer class}
The Categorizer class can be used for pre-processing numerical variables and turning them into categorical variables using a K-Means algorithm. This is essential for the SLBT which requires all variables to be categorical.

\subsubsection{Constructor Parameters}
\begin{itemize}
    \item \textit{method}: specifies the method used to evaluate the optimal number of bins. The allowed values are "elbow", "silhouette" and "fixed". The first uses the elbow method for the sum of squares within the cluster. The second uses the silhouette method. The third forces a specific number of bins. The default, if not given, is "elbow".
    \item \textit{k}: the number of bins into which the numerical variable is categorized. It is necessary when the method is "fixed" and is ignored for any other value. 
    \item \textit{k\_min} and \textit{k\_max}: they specify, respectively, the lowest and highest number of bins possible when the method is "silhouette" or "elbow". The defaults, if not provided, are 2 and 6.
    \item \textit{min\_size}: the minimal number of samples that a bin must contain. If a bin contains fewer samples, it is joined with the closest bin. The default, if not provided, is 5.
    \item \textit{labels}: optional list of names to be assigned to each bin. It must be given as a list of strings of length equal to k. If not provided, the default uses a list on numerical labels (0, 1, 2...). 
\end{itemize}

\subsubsection{Fitting methods}
Once the class has been initialized, there are different methods available in the library to fit and transform the data.

\begin{itemize}
    \item \textit{fit(X)}: learns the binning strategy from the the data.
    \item \textit{transform(X)}: follows the fit method. It transforms the data after the binning strategy has been learned.
    \item \textit{fit\_transform(X)}: allows fitting and transforming data with one single command.
\end{itemize}

With "X" a pandas.DataFrame, pandas.Series, or a numpy.ndarray of the variables to be categorized.

\subsubsection{Reporting method}
With \textit{get\_bin\_info(Column)} it is possible to access a dictionary containing the following information:

\begin{itemize}
    \item \textit{k}: the number of classes.
    \item \textit{centers}: the position of each centroid.
    \item \textit{bins}: the edges of each bin.
    \item \textit{labels}: the label associated with each bin.
\end{itemize}

With "Column" the column name of one of the categorized variables, as a string, or None. If None is provided, the resulting dictionary will contain a report for each categorized variable.

\section{The Pseudo-code algorithms of SLBT}
\label{app:alg}

The pseudocodes of the algorithms for SLBT tree growth and for visual pruning and decision tree selection are provided below.

\begin{algorithm}[H]
\caption{SLBT growth}
\label{alg:slbt-growth}
\KwIn{$\{(Y_i,\bm{X}_i,Z_i)\}_{i=1}^n$, $Z\in\mathcal{G}$; stopping settings; homogeneity type $\mathcal{H}\in\{\text{none}, \text{A}, \text{B}, \text{AB}\}$}
\KwOut{An expanded tree $\mathcal{T}$ with node-wise split information}
\tcp{Step 0: model choice}
Based on the given value of $\mathcal{H}$ select the SLBT variant (U / A / B / AB), defining homogeneity constraints in the SLBM\;
Initialize the tree $\mathcal{T}$ with root node $\{1\}$ containing all observations\;
Initialize the active set $\mathcal{A}\leftarrow\{1\}$\;
\While{$\mathcal{A}\neq\emptyset$}{
    Select a node $t\in\mathcal{A}$ and remove it from $\mathcal{A}$\;
    \tcp{Structural stopping: growth constraints (depth / size / purity)}
    \lIf{StoppingRules$(t)$}{Mark $t$ as leaf and \textbf{continue}}
    \tcp{Step 1: Ranking of predictors}
    \For{$m \gets 1$ \KwTo $M$}{
        Set $\mathrm{GPI}_m(t)\leftarrow \tau_{Y|X_m,Z}(t)$  (Eq.~\eqref{eq:taup})\;
    }
    Sort predictors so that $\mathrm{GPI}_1(t)\ge\dots\ge \mathrm{GPI}_M(t)$\;
    \tcp{Step 2: Model-based split rule}
    $\mathrm{bestPI}\leftarrow -\infty$, $\mathrm{bestSplit}\leftarrow \emptyset$, $\ell\leftarrow 1$\;
    \While{$\ell \le M$}{
        $X \leftarrow X_{(\ell)}$ \;
        Fit SLBM with $K=2$ using predictor $X$  (Eq.~\eqref{eq:SLBM})\;
        Induce split $s_\ell(t)$ via $\hat\alpha_{1|i(g)}(t)\ge 0.5$ (left) vs. $<0.5$ (right)\;
        Compute $\tau_{Y|s_\ell,Z}(t)$\;
        \tcp{Step 3: Split evaluation}
        \If{$\tau_{Y|s_\ell,Z}(t)>\mathrm{bestPI}$}{
            $\mathrm{bestPI}\leftarrow \tau_{Y|s_\ell,Z}(t)$; $\mathrm{bestSplit}\leftarrow s_\ell(t)$
            \;
            $\mathrm{bestPred}\leftarrow X$; $\mathrm{bestGPI}\leftarrow \mathrm{GPI}_\ell(t)$\;
        }
        \tcp{Conceptual early-stop}
        \lIf{$\ell < M$ \textbf{and} $\mathrm{bestPI}\ge \mathrm{GPI}_{\ell+1}(t)$}{\textbf{break}}
        $\ell \leftarrow \ell + 1$\;
    }
    \tcp{Data-driven stopping: no admissible or informative split}
    \lIf{$\mathrm{bestSplit}=\emptyset$}{Mark $t$ as leaf}
    \Else{
        Split node $t$ into children $\{2t\}$ and $\{2t+1\}$ using $\mathrm{bestSplit}$\;
        \tcp{Step 4: Interpretation aids}
        \ForEach{group $g\in\mathcal{G}$ and class $j$}{
            Compute $\mathrm{Lift}_{j|k(g)}(t)$ (Eq.~\eqref{eq:lift}), $k\in\{1,2\}$\;
        }
        Store at node $t$: $\mathrm{bestPred}$, $\mathrm{bestPI}$, $\mathrm{bestGPI}$, $\mathrm{bestSplit}$, $\mathrm{Lift}_{j|k(g)}(t)$\;
        $\mathcal{A}\leftarrow \mathcal{A}\cup\{2t,\,2t+1\}$ \tcp*[f]{add children to active set}\;
    }
}
\Return{$\mathcal{T}$}\;
\end{algorithm}

\begin{algorithm}[H]
\caption{SLBT visual tree selection with leaves interpretation summaries}
\label{alg:slbt-aids-pruning}

\KwIn{A grown tree $\mathcal{T}$; threshold $\varepsilon$}
\KwOut{A selected/pruned tree $\mathcal{T}_{\nu}$ and leaves interpretation summaries}

\tcp{Step 5: Visual pruning and decision tree selection}
\ForEach{node $t \in \mathcal{T}$}{
    Compute $V_Y(t)$ (Eq.~\eqref{eq:nodeIR})\;
    Let $\mathcal{T}_\nu$ be the subtree of $\mathcal{T}$ pruned at $t$\;
    Compute $V_Y(\mathcal{T}_\nu)$ (Eq.~\eqref{eq:treeIR})\;
}

Select the smallest tree $\mathcal{T}_\nu$ such that
$\Delta(\mathcal{T}_\nu)\le \varepsilon$ (Eq.~\eqref{eq:treeDelta})\;

\tcp{Step 4: Interpretation aids}
\ForEach{leaf $h\in{\mathcal{T_\nu}}$}{
    \ForEach{group $g\in\mathcal{G}$ and class $j$}{
        Compute $\mathrm{LCR}_{j|(g)}(h)$ (Eq.~\eqref{eq:LCR})\;
    }
}

\Return{$\mathcal{T}_{\nu}$}\;
\end{algorithm}

\end{document}